\documentclass[lettersize,journal]{IEEEtran}
\usepackage{amsmath,amsfonts}
\usepackage{algorithmic}
\usepackage{algorithm}
\usepackage{array}
\usepackage{subcaption}
\usepackage[caption=false,font=normalsize,labelfont=sf,textfont=sf]{subfig}
\usepackage{textcomp}
\usepackage{textcase}
\usepackage{stfloats}
\usepackage{booktabs}
\usepackage{multirow}
\usepackage[capitalise]{cleveref}
\usepackage{url}
\usepackage{verbatim}
\usepackage{caption}
\usepackage{graphicx}
\usepackage{cite}
\begin{document}

\title{Micro-AU CLIP: Fine-Grained Contrastive Learning from Local Independence to Global Dependency for Micro-Expression Action Unit Detection}

\author{Jinsheng Wei, Fengzhou Guo, Yante Li, Haoyu Chen, Guanming Lu, Guoying Zhao

~\IEEEmembership{Nanjing university of posts and telecommunications$^{1,2,5}$, Center for Machine Vision and Signal Analysis, University of Oulu$^{3,4,6}$}}

\markboth{Journal of \LaTeX\ Class Files,~Vol.~14, No.~8, August~2021}%
{Shell \MakeLowercase{\textit{et al.}}: A Sample Article Using IEEEtran.cls for IEEE Journals}


\maketitle

\begin{abstract}
Micro-expression (ME) action units (Micro-AUs) provide objective clues for fine-grained genuine emotion analysis. Most existing Micro-AU detection methods learn AU features from the whole facial image/video, which conflicts with the inherent locality of AU, resulting in insufficient perception of AU regions. In fact, each AU independently corresponds to specific localized facial muscle movements (local independence), while there is an inherent dependency between some AUs under specific emotional states (global dependency). Thus, this paper explores the effectiveness of the independence-to-dependency pattern and proposes a novel micro-AU detection framework, micro-AU CLIP, that uniquely decomposes the AU detection process into local semantic independence modeling (LSI) and global semantic dependency (GSD) modeling. In LSI, Patch Token Attention (PTA) is designed, mapping several local features within the AU region to the same feature space; In GSD, Global Dependency Attention (GDA) and Global Dependency Loss (GDLoss) are presented to model the global dependency relationships between different AUs, thereby enhancing each AU feature. Furthermore, considering CLIP's native limitations in micro-semantic alignment, a micro-AU contrastive loss (MiAUCL) is designed to learn AU features by a fine-grained alignment of visual and text features. Also, Micro-AU CLIP is effectively applied to ME recognition in an emotion-label-free way. The experimental results demonstrate that Micro-AU CLIP can fully learn fine-grained micro-AU features, achieving state-of-the-art performance.
\end{abstract}


\section{Introduction}
\label{sec:intro}

\IEEEPARstart{M}{icro}-expressions (MEs), as fleeting and involuntary facial expressions, reveal genuine emotions that individuals may attempt to suppress or conceal. As MEs reflect true emotions, ME analysis plays an important role and significance in lie detection, criminal investigation, deep-level human-computer interaction, etc \cite{auflem2022facing,frank1997ability}. At present, ME analysis mainly focuses on ME recognition (MER) that predicts the emotion label of MEs \cite{wei2022learning,fan2023selfme,wei2023prior}. However, ME emotion labels are subjective, and different annotators may label different types of emotions for the same ME sample. The facial action unit (AU) objectively encodes the action state of facial local regions, and research has shown that AU detection can promote fine-grained expression analysis~\cite{Zhou2025Objective}. By detecting micro-expression AU (Micro AU), MEs can be objectively described, avoiding the ambiguity of subjective emotional labels in MER.

\begin{figure}[t]
\centering
\includegraphics[width=0.95\columnwidth]{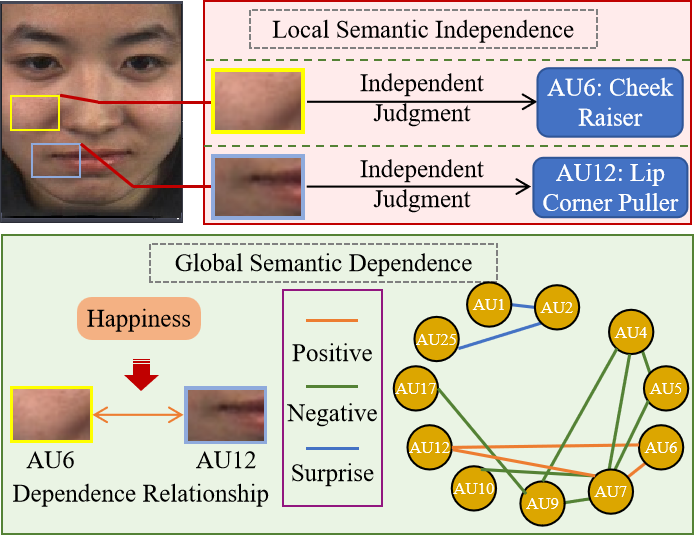} 
\caption{The illustration of the local semantic independence and global semantic dependence.}
\label{motivation}
\end{figure}

Macro-expression (MaE) AU detection typically detects AUs from the entire face, achieving good performance \cite{Shao_Liu_Cai_Ma_2021,li2018eac,shao2019facial}. In micro-AU detection, many works \cite{Zhou2025Objective,li2021intra} have also adopted this pattern to detect micro-AUs and achieved certain results. However, this pattern may hinder the further advancement of micro-AU detection. The primary reason is that the actions encoded by AUs in MaEs are more pronounced, allowing detection models to adaptively capture AU movement information, whereas the subtle actions encoded by micro-AUs in MEs are inherently weak and challenging to detect~\cite{huang2016spontaneous}. Therefore, the detection model is difficult to adaptively capture the micro-AU movement information from the whole facial image/video. 
 Some works \cite{zhou2023micro,zhang2021facial} have also attempted to learn micro-AU features from a local perspective. However, these works have not formed an explicit independent pipeline to focus accurately on each micro-AU region, failing to achieve a fine-grained micro-AU detection.

 In fact, on the one side, AUs have local semantic independence, namely, each AUs can be distinguished individually without the other AUs' information~\cite{Ekman_Friesen_1978}. If an AU is activated, the specific muscle group corresponding to this AU will move, and this movement information is sufficient to represent this AU without other AU activation information~\cite{Zhou2025Objective}. As shown in \cref{motivation}, peoples only need to observe the movement information of the lip corner area to determine whether AU12 is activated, without needing to check the movement information of other AU-specific areas, such as the cheek. On the other side, certain AUs exhibit global semantic dependencies, that is, specific true emotions trigger the co-occurrence of AU groups in MEs~\cite{Ekman_Friesen_1978}. \cref{motivation} shows that AU12 and AU6 usually exhibit co-activation in a "Happiness" ME instance. If AU12 is activated, even if the confidence level of AU6 being activated is around 0.5 (borderline confidence), we can still infer that the probability of AU6 being activated should be positive~\cite{Zhou2025Objective}. Subtle facial movements make micro-AU detection challenging~\cite{huang2016spontaneous}. Based on the above theory, we argue that modeling local semantic independence and global semantic dependence can alleviate this challenge to enhance micro-AU detection performance.
Therefore, this paper proposes a fine-grained micro-AU detection framework in a "from local independence to global dependence ( independence-to-dependence )" pattern and explores the effectiveness of this pattern.

Inspired by the excellent visual-text alignment ability of Contrastive Language-Image Pre-training (CLIP)~\cite{radford2021learning}, this paper instantiates the above framework based on CLIP and proposes a novel Micro-AU CLIP model that designs fine-grained text to guide the visual encoder to learn fine-grained micro-AU features. Micro-AU CLIP model includes a local semantic independence (LSI) and a global semantic dependency (GSD) module. LSI independently infers the movement features of each AU, and GSD learns the global dependency between all AU features. However, conventional CLIP excels at macro semantic alignment and has limitations in fine-grained semantic alignment~\cite{zhang2025clip}. Therefore, we design a micro-AU contrast loss (MiAUCL) for the local fine-grained alignment between AU visual and text semantics. MiAUCL can guide the model to learn a single fine-grained AU visual feature by utilizing the textual description of the individual AU.

Furthermore, as is well known~\cite{wei2023geometric}, there is a strong and clear correlation between micro-AU and facial MEs. Thus, the AU detection results can be used for ME recognition (MER). However, directly inferring ME categories based on discrete AU predicted labels may lead to accumulated errors and fail to achieve competitive performance. Some works \cite{wang2024micro,Zhou2025Objective} introduce AU information to improve the MER performance, but these works still need ME label to achieve MER tasks through supervised learning or weakly supervised learning. Differently, our work combines Micro-AU CLIP with emotion-label-free methods to achieve MER. To our knowledge, this is the first work that applied emotion-label-free to MER, based on AU-supervised transfer.

Thus, the main contributions of this paper are as follows:

\begin{itemize}

    \item This work explores a novel Micro-AU detection paradigm from local independence to global dependence learning by transforming isolated local estimation into a unified modeling of both local independence and global dependency for effective AU modeling. Micro-AU CLIP model can be applied to MER, showing strong generalization beyond supervised settings.


    \item In LSI, Patch Token Attention (PTA) is designed to learn AU semantic consistency features within the corresponding AU region. 
    \item In GSD, Global Dependency Attention (GDA) with global dependency loss is designed to build the global dependency relationship between AUs to enhance micro-AU features. 
    \item To achieve fine-grained cross-modal alignment, MiAUCL is proposed to achieve fine-grained guidance of text information on visual micro-AU feature learning.



\end{itemize}

\section{Related Works}
Micro-expression analysis involves multiple tightly related research topics, including Micro-AU detection, model learning strategies and MER. Existing works have explored these aspects from different perspectives, such as global or local feature modeling, the idea of contrastive learning and various supervised or weakly supervised learning paradigms. In this section, we review the most relevant literature from three directions: Micro-AU detection paradigm, which focuses on extracting discriminative AU representations from subtle facial motions; Contrastive learning for ME analysis, which emphasizes visual–text or intra-/inter-sample alignment; and MER learning ways, which summarize mainstream learning paradigms for emotion recognition. The related works highlights the limitations of existing methods and motivates the design of our Micro-AU CLIP framework.
\subsection{Micro-AU Detection Paradigm}
Micro-AU detection aims to recognize subtle facial muscle movements associated with micro-expressions, and has attracted increasing attention in recent years. Many studies \cite{varanka2023learnable,li2023con,zhou2023micro} mainly attempted to capture micro-AU movement information from the entire facial ME images or video sequences. For example, Liu et al.~\cite{liu2025llm} extracted intermediate and high-level visual features from the 3D-CNN backbone across the entire face and fuse them for AU detection. Tuomas et al.~\cite{varanka2023learnable} extracted global spatio-temporal information from video sequences via 3D convolution in order to capture the global trend of MEs. 

Although global modeling strategies can capture holistic motion patterns, they often overlook the fact that AUs are inherently localized and correspond to specific facial muscle groups. To address this issue, several studies focus on feature extraction in local regions. To extract local AU features, Li et al.~\cite{Li_Huang_Zhao_2021} propose the Spatio-Channel Attention (SCA) mechanism to model intra- and inter-region information interactions, enabling the model to emphasize AU-related regions. However, these works tend to entangle multiple AUs into holistic features, weakening AU-level discriminability and ignore the independence of local micro-AUs. Zhang et al.~\cite{zhang2021facial} use facial marker points to segment the image into 11 key regions, and extract AU features of each region, making independent AU predictions for each region. Although they make independent judgments for each AU, they overlook the global dependencies of AUs and the structured relationships among AUs.

Different from these methods, this paper designs a unified framework that follows a from local independence to global dependence paradigm. Specifically, each AU feature is first independently learned through explicitly separated pipelines to preserve local semantic independence, and then global AU dependencies are modeled by a dedicated module, balancing local independence and global dependency, thus achieving the integration of local and global features.

\begin{figure*}[t]
\centering
\includegraphics[width=1\textwidth]{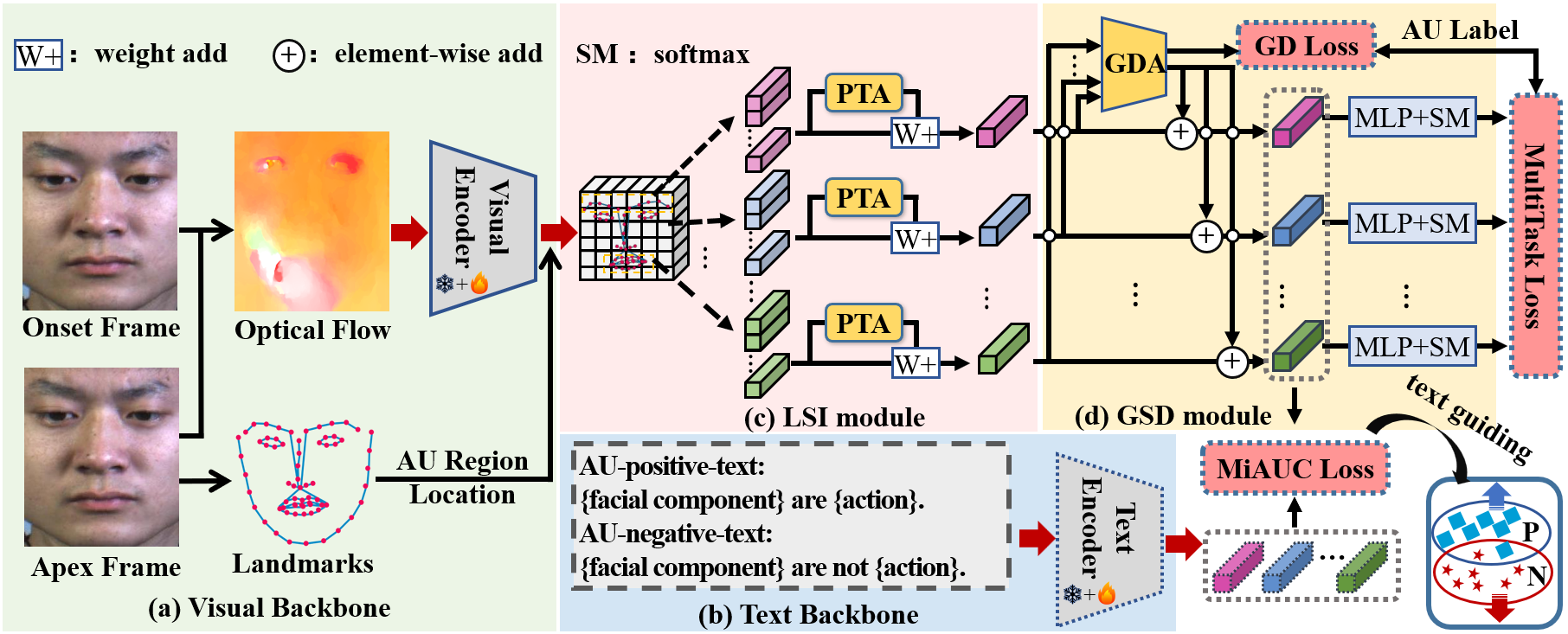} 
\caption{Framework of the proposed Micro-AU CLIP. (a) ViT-based visual backbone extracts patch-level visual tokens from the input image.; (b) Text backbone encodes AU-related textual prompts into text features; (c) In LSI, the model utilizes PTA to learn each AU feature individually; (d) GSD models the global dependence between different AUs and propagates the global context information to the features of each AU.}
\label{figframework}
\end{figure*}
\subsection{Contrastive Learning}
Contrastive learning has demonstrated remarkable effectiveness in multimodal representation learning and vision–language alignment, and has recently been introduced into fine-grained vision tasks. In the context of ME analysis, Li et al.~\cite{li2021intra} proposed an Intra- and Inter-Contrastive Learning (ICL) strategy to enlarge the feature differences between onset and apex frames, as well as between different AUs, thereby enhancing motion-sensitive features. Based on CLIP, Liu et al.~\cite{liu2025mer} perform alignment between global ME visual features and emotion-level text descriptions, enabling cross-modal contrastive learning. In other fields, such as multimodal sentiment analysis and generative artificial intelligence, research has also applied the contrastive learning idea based on CLIP. Qi et al.~\cite{qi2024multimodal} fine-tuned CLIP for the multimodal sentiment recognition task, aligning video frame features with emotional text descriptions semantically, in order to extract video features with stronger sentiment discriminability. Guo et al similarly aligned image features with text features, achieving controllable guidance from text semantics to video generation.

Although existing contrastive learning methods have achieved success in ME analysis, they primarily operate at the global level and cannot effectively distinguish fine-grained AU semantics from a local perspective. Furthermore, standard CLIP-style contrastive learning was originally designed for single-label single-label classification tasks and is not suitable for multi-label AU detection tasks.

Different from these methods, we design a novel MiAUCL loss to achieve fine-grained alignment between local visual features and text semantics for each AU. MiAUCL does not perform contrastive learning on global features, but instead performs visual-language alignment for each AU, allowing each local AU visual feature to be guided by its corresponding semantic concept. This fine-grained alignment strategy provides effective text-level supervision for local visual features and is more in line with the multi-label characteristics of AU detection.

\subsection{MER Learning Way}
So far, the dominant learning approaches of most existing MER methods \cite{gong2023meta,cai2024mfdan} are supervised learning in MER. However, ME datasets are typically small-scale, and emotion annotations are often ambiguous and subjective, some works \cite{qi2024multimodal,nguyen2023micron} attempt to implement MER tasks using weakly supervised learning methods. For instance, Almushrafy et al. achieved efficient intensity estimation solely through sparse temporal labeling of micro-expressions, without relying on frame-by-frame intensity annotation. Nguyen et al.~\cite{nguyen2023micron} pre-train models on large-scale unlabeled data by constructing self-supervised tasks to learn general ME features. While these approaches reduce dependence on fully labeled data, they still require emotion labels or pseudo-labels during training.

Different from these methods, based on the proposed Micro-AU CLIP, we exploit the emotion-label-free MER, mitigating the reliance on ME emotion labels. This provides a new perspective for MER, mitigating reliance on explicit emotion annotations and enabling flexible transfer to downstream emotion recognition tasks.

\section{Proposed Method}
\cref{figframework} presents the framework of Micro-AU CLIP model. First, the visual and text backbones extract visual features and text features, respectively; second, LSI learn micro-AU independence features using patch token attention (PTA) in an independent pipeline pattern; next, GSD module models the dependence relationship between AUs feature using global dependence attention (GDA) with global dependence loss (GDL); then,  MiAUCL enhances the positive (P) and negative (N) sample differences for each AU, and Multi-Task Loss constrain the classifier to detect the activation status of each AU. Finally, Micro-AU CLIP model is employed to achieve emotion-label-free MER.

\subsection{Visual and Text Backbone}
CLIP has powerful visual and text alignment capabilities and is effectively applied in some image analysis tasks~\cite{zhao2023clip,liu2025mer}. Micro-AU CLIP employed CLIP's visual and text encoders as the backbone networks for visual and text feature extractions, respectively.

Optical flow (OF) can represent the motion information between two frames, which is consistent with the dynamic nature of micro-AUs. OF between the onset and apex frames has been widely used in MER \cite{Zhou2025Objective,wang2024two,wang2024micro}. Thus, the proposed method input OF to a visual encoder (VE) with pre-trained weights, extracting the visual token $F^V_O$ $\in \mathbb{R}^{  h\times w \times d}$, where $d$ is the feature dimension; $h$ and $w$ represent the token number in the horizontal and vertical directions, respectively.

To learn fine-grained micro-AU features, the proposed method sets the text descriptions of positive (P) and negative (N) samples for each AU, as shown in \cref{figframework}. 
For example, for AU4, the description of P samples is "The eyebrows are lowering"; that of N samples is “The eyebrows are not lowering”. 
Assuming there are $N$ AUs, each sample obtains $N$ text descriptions $T_n (n=1, 2, ..., N)$. $T_n$ is processed by the text encoder to obtain AU text features $F^T_n$.

\subsection{LSI Module}
Micro-AUs have local semantic independence, and in the feature learning stage, independent feature learning for each AU can avoid semantic interference between different AUs~\cite{zhang2021facial}. Therefore, we design an LSI module to learn fine-grained micro-AU features, including two parts: fine-grained local feature partitioning and patch token attention.

\subsubsection{Fine-Grained Local Feature Partitioning}
AU activates specific facial muscle movements, and facial landmarks can locate these specific muscle regions~\cite{wei2025multi}. Furthermore, there is spatial position consistency between the patch token of the visual encoder and the input OF. Thus, to partition the motion features corresponding to each AU region, facial landmarks are extracted to partition $F^V_O$ features into multiple local AU feature groups. 

\begin{figure}
    \centering
    \includegraphics[width=0.8\linewidth]{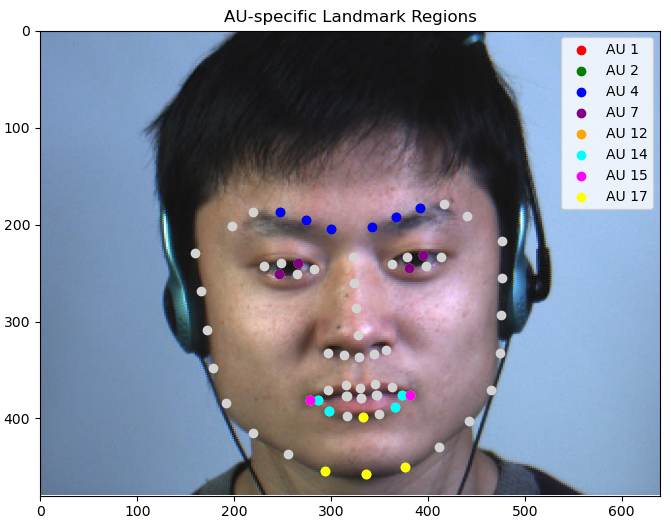}
    \caption{Illustration of AU and the corresponding facial landmarks.}
    \label{fig:68}
\end{figure}

68 facial landmarks are extracted. The mapping relationship between AUs and the corresponding landmarks is shown in \cref{landmarks} and \cref{fig:68}, based on the FACS (Facial Action Coding System). Specifically, AU1, AU2, and AU4 all involve the eyebrow region, so the landmarks in the middle of the eyebrows (19–24) were selected; AU7 corresponds to lower eyelid contraction, so keypoints below the eyes (38, 41, 44, 47) were selected; AU12 and AU15 affect the upward or downward movement of the corners of the mouth, so points in the corner of the mouth region (48, 54) were jointly selected; AU14 affects the outward pulling of the lips, involving points in the middle and sides of the mouth (55, 59, 60, 64); AU17 represents chin contraction or lip upward movement, so points at the center of the chin (7, 8, 9, 57) were selected. This landmark selection strategy, based on the correspondence between anatomical regions and landmarks, enables the model to perform more precise region localisation and feature extraction for each AU's key region, thereby enhancing the spatial sensitivity and interpretability of AU detection.

\begin{table}[h]
    \centering
    \caption{\textsc{The mapping between AUs and facial landmarks}}
    \label{landmarks}
    \begin{tabular}{|c||c|}
        \hline
        \textbf{AU} & \textbf{The index number of facial landmarks} \\
        \hline
        1  & 19, 20, 21, 22, 23, 24 \\
        \hline
        2  & 19, 20, 21, 22, 23, 24 \\
        \hline
        4  & 19, 20, 21, 22, 23, 24 \\
        \hline
        7  & 38, 41, 44, 47 \\
        \hline
        12 & 48, 54 \\
        \hline
        14 & 55, 59, 60, 64 \\
        \hline
        15 & 48, 54 \\
        \hline
        17 & 7, 8, 9, 57 \\
        \hline
    \end{tabular}
\end{table}

To obtain the location information of AUs, the landmarks are extracted from the apex frame. Assuming that the facial area corresponding to the $n$-th AU includes $N_L$ facial landmarks $L^n_j, j=1,2,...,N_L$. Given a landmark $L^n_j (x^n_j,y^n_j)$ , we locate the corresponding patch token $K^V_{n,j}$ in $F^V_O$ based on the coordinate $(x^n_j,y^n_j)$ of $L^n_j$ as follows:

\begin{equation}
   K^V_{n,j}=F^V_O (x^n_j \times \frac{h}{h_0}, y^n_j \times \frac{w}{w_0},:) \in \mathbb{R}^{  d},
\end{equation}
where $h_0$ and $w_0$ are the height and width of OF, namely those of apex frame, respectively. Thus, for each AU, we can obtain a group of patch tokens \{$K^V_{n,j}\} (j=1,2,...,N_L)$.

\subsubsection{Patch Token Attention}
\begin{figure}[t]
    \centering
    \resizebox{0.5\textwidth}{!}{%
    \begin{tabular}{c} 
        \begin{subfigure}[b]{\linewidth} 
            \centering
            \includegraphics[width=\linewidth]{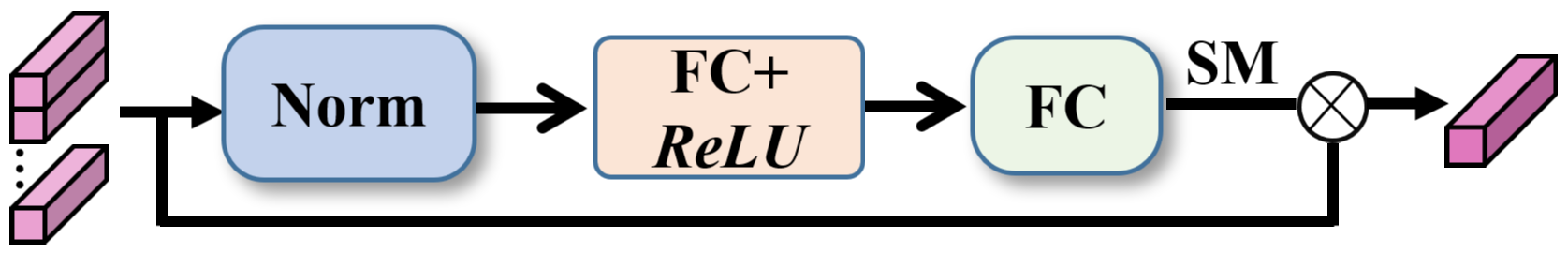}
            \caption{\small Patch Token Attention}
            \label{atten-a}
        \end{subfigure} \\ 
        \vspace{0.3cm} 
        \begin{subfigure}[b]{\linewidth}
            \centering
            \includegraphics[width=\linewidth]{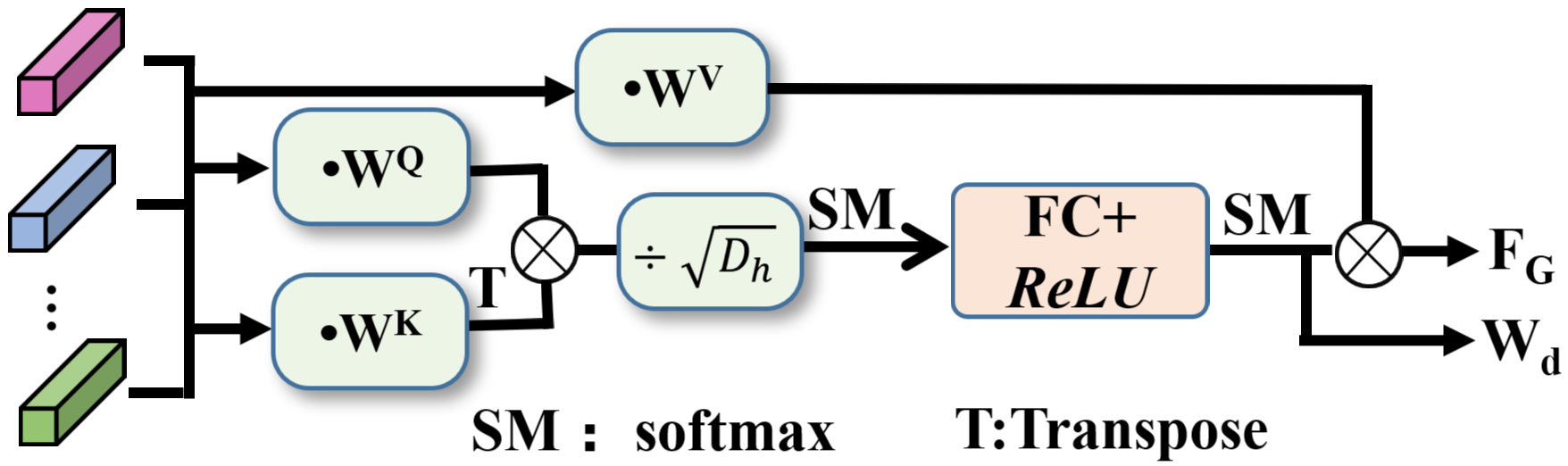}
            \caption{\small Global Dependency Attention}
            \label{atten-b}
        \end{subfigure}
    \end{tabular}
    }
    \caption{The details of PTA and GDA.}
    \label{atten}
\end{figure}

A group of patch tokens corresponding to a specific AU has local semantic consistency information. However, the contribution of each patch token to micro-AU detection varies.
For example, AU1 represents the raising of the inner eyebrow. 
That is, the patch token corresponding to the inner eyebrow is definitely more important than the patch token corresponding to the outer eyebrow. Thus, as shown in \cref{atten-a}, this paper designs a PTA that weights and fuses all patch tokens within the $n$-th AU region, based on their contributions.

For $n$-th AU, we first learn a contribution weight $W^n_j$ for $j$-th patch token $K^V_{n,j}$:
\begin{equation}
W^n_j=\frac{e^{\mathrm{MLP}(K^V_{n,j})}}{\sum_{j=1}^{N_L} e^{\mathrm{MLP}(K^V_{n,j})}} ,
\end{equation}
where MLP is Multiple Layer Perceptron that map $d$-dimensional vector $K^V_{n,j}$ to a scalar. Then, the learned weights \{$W^n_j$\} are used to the weighted sum for the patch tokens \{$K^V_{n,j}$\} to obtain local micro-AU features $\tilde{F}^V_n$:

\begin{equation}
\tilde{F}^V_n=\sum_{j=1}^{N_L} W^n_j K^V_{n,j} \in \mathbb{R}^{  d}, 
\end{equation}

\subsection{GSD Module}

Under specific emotions, certain facial action units (AUs) tend to appear synergistically, indicating a global semantic dependency between them. The Bayesian criterion provides~\cite{van2021bayesian} a theoretical basis for modeling this dependency relationship: given specific emotional conditions, if there is a conditional dependency relationship between AUs, observing the state of some AUs can help infer the probability of other AUs appearing~\cite{Zhou2025Objective}. Based on this principle, this paper designs a GSD module aimed at learning and establishing the global dependency relationship between AUs.

In GSD module, Global Dependency Attention models the global dependency relationship of micro-AU feature $\tilde{F}^V_n$ to learning global dependency features that then are introduced with each micro-AU feature $\tilde{F}^V_n$. Also, the labels of the co occurring AUs in the ME samples are used to constrain the attention weights learned by GDA, for focusing on the AU features with dependency relationship, as shown in \cref{atten-b}.

\subsubsection{Global Dependency Attention}

Inspired by the self-attention mechanism, we first learn the self-attention weights between different AU features and perform nonlinear mapping to learn the dependency weights between different AU features. First, self-attention weights are calculated by:
\begin{equation}
    A=\text{softmax}\left(\frac{QK^T}{\sqrt{D_h}}\right)
    \label{eq:placeholder_label}
\end{equation},
where $Q=\tilde{F}^VW^Q$, $K=\tilde{F}^VW^K$, and $Q$, $ K \in \mathbb{R}^{d \times d}$; $\tilde{F}^V=\{\tilde{F}^V_1,\tilde{F}^V_2,...,\tilde{F}^V_N\}\in \mathbb{R}^{N \times d} $. $W^Q,W^K$ are the learnable linear projection matrices.

Then, we apply a fully connected layer $\mathrm{FC}$ and softmax layer ($Softmax$) to obtain the dependency weights:
\begin{equation}
    W_d = Softmax(ReLU(\mathrm{FC}(A))\in \mathbb{R}^{1 \times N}
\end{equation}

Next, based on the dependency weight $W_d$, the global dependency feature $F_G$ is obtained by:
\begin{equation}
    F_G = W_d \tilde{F}^V W^V\in \mathbb{R}^{1\times d},
    \label{eq:placeholder_label}
\end{equation}
where $W^V$ is a linear projection matrix. Finally, the global dependency information is introduced to each AU into  obtain the probability of $n$-th AU being activated or not by:

\begin{equation}
P_n=Softmax(\mathrm{MLP}(F_G+\tilde{F}^V_n)) \in \mathbb{R}^{2}, 
\end{equation}
where, $F_G+\tilde{F}^V_n$ is the final $n$-th AU visual feature. 

\subsubsection{Global Dependency Attention Loss}
To avoid interference from AU features without dependency relationships, we designed the global dependency loss to ensure consistency between the dependency weights and the AUs that have the dependency  relationship. Assuming the label of N AUs is $Y \in \mathbb{R}^{1\times N}$, then GDL loss is defined as:
\begin{equation}
    \mathcal{L}_{\text{GD}} = \frac{1}{N}  \left\| W_d - Y \right\|_2^2
    \label{eq:placeholder_label}
\end{equation}

It is worth noting that although the dependency weights are supervised by the real labels of AU, the proposed GSD module dependencies at the feature interaction level rather than the label level. Dependency weights are generated based on the interactions between the features of each action unit, and determine how each AU feature aggregates information from other action unit features, thereby constructing a clear information flow from action unit to action unit. Therefore, the model can still learn AU-AU conditional dependencies or co-occurrence structures.

\subsection{MiAUC Loss}

CLIP can align visual and text features, and enhance the consistency of positive sample pairs and the difference of negative sample pairs through contrastive loss~\cite{radford2021learning}. However, the CLIP model has limitations in fine-grained tasks~\cite{zhang2025clip}. Therefore, this paper proposes a fine-grained Micro-AU Contrastive Loss (MiAUCL), which utilizes text features to guide the model in learning more fine-grained visual micro-AU features.

MiAUC Loss align visual and textual features separately for each AU, explicitly guiding them at a fine-grained level. First, for $n$-th AU, given a training batch with $B$ samples, $B$ visual AU features $\{F^V_n\}$ and $B$ text features $\{F^T_n\}$ can be obtained. Then, we can obtain the cosine similarity matrix:

\begin{equation}
M_n=\mathrm{CS}([F^V_{n,1};...;F^V_{n,B}],[(F^T_{n,1};...;F^T_{n,B}]), 
\end{equation}
where $F^V_{n,b}$ and $F^T_{n,b}$ express the visual and text AU feature of $b$-th sample in a batch for $n$-th AU, respectively; $M_n \in \mathbb{R}^{B\times B}$; CS is the cosine similarity function.

In CLIP, only the diagonal matrix in the similarity label is set to 1~\cite{radford2021learning}. However, each AU has only two categories, namely activated and inactivated, and simply using diagonal matrices as label matrices can easily overlook the similarity between different samples with the same category. If the $b1$-th and $b2$-th samples have the same category for $n$-th AU, the similarity between $F^V_{n,b1}$ and $F^T_{n,b2}$ should be 1. Thus, the label matrix $LM_n$ of MiAUC loss is:
\begin{equation}
LM_n(b1,b2)=
\begin{cases}
1,& {\text{if}}\ C(F^V_{n,b1})= C(F^T_{n,b2})\\
0,& {\text{otherwise.}} 
\end{cases}
\end{equation}
, where $C(X)$ expresses the AU label corresponding to X; $b1$ and $b2$ are from 1 to $B$. 

Finally, cross-entropy loss is used for $LM_n$ and $M_n$ to obtain the contrastive loss $\text{CL}_n$ of the $n$-th AU, and then the contrastive losses of all AUs are added together to obtain the final MiAUC loss:
\begin{equation}
L_{\text{MiAUC}} =\sum ^{N} _{n=1}\text{CL}_n.
\end{equation}

As a result, the total loss is calculated by:
\begin{equation}
\mathrm{Loss} = \mathrm{MultiTask \,  Loss } + \alpha \mathcal{L}_{MiAUC} + \beta \mathcal{L}_{\mathrm{GD}} ,
\end{equation}
where $\alpha$ and $\beta$ is the trade-off coefficient. MultiTaskLoss is a multi-label AU classification loss, which is implemented as the sum of cross-entropy losses for each AU.

\subsection{Emotion-Label-Free MER}

There is a strong correlation between ME emotions and micro-AU. MEs can be identified by the activation status of AU~\cite{Zhou2025Objective}. Directly using AU detection results for MER can easily lead to information loss due to AU discretization.
This paper calculates the similarity between the visual features $F^V_n$ and label text features $LF^T_r$ of each AU and sums them up to determine the emotional category of MEs:

\begin{equation}
\mathrm{MaxME}({\sum ^{N} _{n=1} \mathrm{CS}(F^V_n,LF^T_r})), r=1,...,R,
\end{equation}
where MaxME(*) expresses the emotion category with max similarity; $R$ is the total number of label texts.


\section{Experiments}
This section reports the experimental results. We study the
effectiveness of local independence to 
global dependence pattern and evaluate the performance of the proposed method. Firstly, the effectiveness of the proposed pattern is studied for micro-AU. Secondly, we carry out the ablation analysis with some visualizations to evaluate the proposed model and components, including PTA, GDA and MiAUC loss. Thirdly, the parameter evaluation is carried out; Finally, we compare the proposed method with state-of-the-art (SOTA) methods.
\subsection{Experiment Setting}

\subsubsection{Datasets and Evaluation Metrics}
 
As following existing works \cite{varanka2023learnable, Zhou2025Objective,Li_Huang_Zhao_2021}, CASME II and SAMM are adopted to evaluate the performance of the proposed method. 
CASME II~\cite{Yan_Li_Wang_Zhao_Liu_Chen_Fu_2014} contains 255 samples with 26 subjects, collected by high-speed cameras with 200 fps. 
SAMM~\cite{davison2016samm} contains 159 samples with 32 subjects, collected by high-speed cameras with 200 fps. 
Consistent with existing works \cite{zhang2021facial,li2021micro}, common AUs with a number of samples greater than 14 are utilized. Thus, eight and four AUs are selected on CASME II and SAMM, respectively. Select AU1, 2, 4, 7, 12, 14, 15, 17 on CASME II; select AU2, 4, 7, 12 on SAMM.
For MER task, MEs are classified into
three categories: positive, negative and surprise, as follows existing works \cite{zhang2022short,liong2018less}.

Following existing mainstream works \cite{varanka2023learnable, Zhou2025Objective,Li_Huang_Zhao_2021}, All experiments were performed under Leave-One-Subject-Out (LOSO) cross-validation. 
The results of all subjects are accumulated to calculate the F1-score (F1) as the evaluation metric. In addition, following the mainstream work on macro-expression AU detection \cite{Shao_Liu_Cai_Ma_2021,shao2019facial}, we add the accuracy (ACC) as an evaluation metric.

\subsubsection{Implementation Setting}
As in existing works~\cite{li2021intra}, the onset and apex frames were obtained using database labels, and their detection belongs to another task in ME analysis. Facial micro-actions are magnified using learning-based amplification~\cite{oh2018learning} with a magnification factor of 3. The Dlib package is employed to detect facial landmarks. The ViT-B/32 visual encoder are employed. The OF image is resized to 224×224 pixels. 
In model training, we fine-tune the last three layers of both the visual and text encoders. The epoch number and batch size are set to 80 and 8, respectively. The model is optimized by an SGD optimizer with a 0.001 learning rate for visual and text encoders, and a 0.01 learning rate for other modules. The learning rate is divided by 10 at the fortieth epoch. Random seed set to 1. All models are trained on a RTX 4060 Ti (16G) with Pytorch 2.6.0 version.

\subsection{The Study on the Effectiveness of Local Independence to Global Dependence Pattern}

\begin{table}[t]
\caption{\textsc{The Ablation Results on LSI and GSD}}
\label{tabframework}
\centering
  \begin{tabular}{|c||c||c||c||c||c|}
    \hline
    \multicolumn{2}{|c||}{Module}&\multicolumn{2}{c||}{CASME II} &\multicolumn{2}{c|}{SAMM}\\
    \cline{1-2} \cline{3-4} \cline{5-6}
    LSI  &GSD &F1 &ACC &F1 &ACC\\
    \hline
    $\checkmark$ &$\times$     &0.761&0.904&0.714&0.846\\
    \hline
    $\times$     &$\checkmark$ &0.730&0.898&0.638&0.812\\
    \hline
    $\checkmark$ &$\checkmark$ &\textbf{0.782}&\textbf{0.917}&\textbf{0.730}&\textbf{0.863}\\
    \hline
  \end{tabular}
\end{table}
This paper explores the effectiveness of the fine-grained Micro-AU detection framework from local independence to global dependence. Thus, we conduct the ablation experiments on LSI and GSD to evaluate the effectiveness of the framework. As shown in \cref{tabframework}, whether removing LSI or GSD, performance has decreased. It turns out that the two components are effective, and this framework can be effectively applied to micro-AU detection tasks.

Furthermore, to verify whether the text encoder requires fine-tuning, we compared two ways: fixed weights and fine-tuning. \cref{tabfinetune} shows that the fine-tuning is superior to fixed weights. Under fixed weights, the text feature similarity of negative sample pairs is close to 1, while that under fine-tuning is close to -1. Thus, fine-tuning is more suitable for our task requirements. More analysis is available in the supplementary materials.

\begin{table}[t]
\caption{\textsc{The Evaluation on the Finetune of the Text-Encoder. TextSi : Similarity between Positive and Negative Text}}
\label{tabfinetune}
\centering
  \begin{tabular}{|c||c||c||c||c||c|}
    \hline
    \multirow{2}{*}{Module}&\multicolumn{2}{c||}{CASME II} 
    &\multicolumn{2}{c||}{SAMM}& \multirow{2}{*}{TextSi}\\
    \cline{2-3} \cline{4-5}
     &F1 &ACC &F1 &ACC&\\
    \hline
    fixed &0.769&0.898&0.702&0.812&0.994\\
    \hline
    finetune &\textbf{0.776}&\textbf{0.912}&\textbf{0.716}&\textbf{0.853}&-0.902\\
    \hline
  \end{tabular}
\end{table}

\subsection{Ablation Study}

\subsubsection{The Evaluation on $\alpha$}

We evaluated the trade-off coefficient $\alpha$ from 0.2 to 1 with a 0.2 interval.
As shown in \cref{tab:alpha}, the optimal $\alpha$ on CASME II dataset is 0.6, and increasing or decreasing it will cause a significant decrease in performance; On SAMM dataset, the optimal $\alpha$ is 1, and the performance is similar at 0.2 and 0.6. This difference may be due to different numbers of AUs. In addition, the differences in sample distribution and collection equipment in the dataset can also cause variations.

\begin{table}[t]
\caption{\textsc{The Ablation Study on $\alpha$}}
\label{tab:alpha}
\centering
  \begin{tabular}{|c||c||c||c||c|}
    \hline
    \multirow{2}{*}{$\alpha$}&\multicolumn{2}{c||}{CASME II} 
    &\multicolumn{2}{c|}{SAMM}\\
    \cline{2-3} \cline{4-5}
     &F1 &ACC &F1 &ACC\\
    \hline

    0.2 &0.770&0.908&0.712&0.852\\
    \hline
    0.4 &\textbf{0.776}&0.908&0.687&0.833\\
    \hline
    0.6 &\textbf{0.776}&\textbf{0.912}&0.715&0.849\\
    \hline
    0.8 &0.764&0.911&0.696&0.849\\
    \hline
    1.0 &0.767&\textbf{0.912}&\textbf{0.716}&\textbf{0.853}\\
    \hline
  \end{tabular}
\end{table}

\subsubsection{The Evaluation on $\beta$}
We evaluated the trade-off coefficient $\beta$ from 0.2 to 1 with a 0.2 interval based on $\alpha$ is set to 0.6 on CASME II and 1 on SAMM. As shown in \cref{tab:beta}, on the CASME II dataset, the model achieves optimal performance at $\beta$ = 1 and the accuracy under all parameters remains between 0.911 and 0.917, demonstrating strong robustness. On the SAMM dataset, the model reached a better performance at $\beta$ = 0.6, with performance generally declining as $\beta$ increased thereafter, demonstrating that SAMM is more sensitive to $\beta$ than $\alpha$.

\begin{table}[t]
\caption{\textsc{The Ablation Study on $\beta$}}
\label{tab:beta}
\centering
  \begin{tabular}{|c||c||c||c||c|}
    \hline
    \multirow{2}{*}{$\beta$}&\multicolumn{2}{c||}{CASME II} 
    &\multicolumn{2}{c|}{SAMM}\\
    \cline{2-3} \cline{4-5}
     &F1 &ACC &F1 &ACC\\
    \hline
    
    0.2 &0.780&0.914&0.704&0.848\\
    \hline
    0.4 &0.775&0.912&0.712&0.852\\
    \hline
    0.6 &0.776&0.914&\textbf{0.730}&\textbf{0.863}\\
    \hline
    0.8 &0.769&0.911&0.693&0.843\\
    \hline
    1.0 &\textbf{0.782}&\textbf{0.917}&0.711&0.855\\
    \hline
  \end{tabular}
\end{table}

\subsubsection{The Evaluation on PTA and GDA}

We evaluated the effectiveness of PTA in LSI. As shown in \cref{tab:PTA}, compared to maxpooling and meanpooling, PTA has better performance. It indicates that weighting the contribution of different patch tokens is effective. Also, meanpooling is worse than maxpooling, indicating that it is not appropriate to consider the contributions of different patch tokens as equal. 

To further evaluate the methods of GDA in GSD module. \cref{tab:GDA} demonstrate that Add+MLP is superior to Cat+MLP. The reason may be that although concatenation preserves all information, it also stacks up potentially unrelated, redundant, and even noisy information together. Subsequent MLP requires a larger model capacity and training data to learn how to ignore irrelevant information and correctly associate features from different locations, which increases the difficulty of learning micro AU features.

\begin{table}[t]
\caption{\textsc{The Ablation Results on PTA}}
\label{tab:PTA}
\centering
  \begin{tabular}{|c||c||c||c||c||c|}
    \hline
    \multirow{2}{*}{Method}&\multirow{2}{*}{Module}&\multicolumn{2}{c||}{CASME II} 
    &\multicolumn{2}{c|}{SAMM}\\
    \cline{3-4} \cline{5-6}
    & &F1 &ACC &F1 &ACC\\
    \hline

    meanpool &LSI&0.754&0.902&0.680&0.837\\
    \hline
    maxpool &LSI &0.765&0.909&0.709&0.849\\
    \hline
    PTA &LSI&\textbf{0.776}&\textbf{0.912}&\textbf{0.716}&\textbf{0.853}\\
    \hline
  \end{tabular}
\end{table}

\begin{table}[t]
\caption{\textsc{The Ablation Results on GDA}}
\label{tab:GDA}
\centering
  \begin{tabular}{|c||c||c||c||c||c|}
    \hline
    \multirow{2}{*}{Method}&\multirow{2}{*}{Module}&\multicolumn{2}{c||}{CASME II} 
    &\multicolumn{2}{c|}{SAMM}\\
    \cline{3-4} \cline{5-6}
    & &F1 &ACC &F1 &ACC\\
    \hline

    Cat+MLP &GSD&0.771&0.909&0.641&0.833\\
    \hline
    Add+MLP &GSD&0.776&0.912&0.716&0.853\\
    \hline
    GDA &GSD&\textbf{0.782}&\textbf{0.917}&\textbf{0.730}&\textbf{0.863}\\
    \hline
  \end{tabular}
\end{table}

\subsubsection{The Evaluation on Fine-Grained Contrastive Loss}
MiAUC Loss plays an important role in fine-grained micro-AU feature learning. We compared it with three other methods: No CL, without contrastive learning, i.e. without the text encoder; global OrigCL, the label matrix of the text encoder is the overall description of all AU; local OrigCL, the label matrix of the text encoder is a diagonal matrix. 

\cref{tab:rMAU-CL} shows the comparison results. Overall, our method achieved the best performance, indicating that the proposed fine-grained contrastive loss is effective for micro AU detection. Furthermore, global OrigCL achieved the worst performance, indicating that it is inappropriate to guide the model to learn global micro-AU features through the coarse-grained AU description. Compared to global OrigCL, local OrigCL has achieved significant advantages, indicating that partitioning global features into individual AU features and applying fine-grained constraints through several local contrastive losses is an effective fine-grained learning approach. Compared to the proposed method, local OrigCL has a slight disadvantage, which also indicates that label vectors based on diagonal matrices ignore the similarity between samples with the same AU category. Interestingly, No CL is superior to global OrigCL, indicating that 1) our framework is also effective without text guidance; 2) using global text descriptions to guide the model in learning micro-AU features did not have a positive effect.

\begin{table}[t]
\caption{\textsc{The Ablation Study on MiAUC Loss}}
\label{tab:rMAU-CL}
\centering
  \begin{tabular}{|c||c||c||c||c|}
    \hline
    \multirow{2}{*}{Loss}&\multicolumn{2}{c||}{CASME II} 
    &\multicolumn{2}{c|}{SAMM}\\
    \cline{2-3} \cline{4-5}
     &F1 &ACC &F1 &ACC\\
    \hline

    No CL &0.759&0.902&0.725&0.853\\
    \hline
    global OrigCL &0.730&0.898&0.638&0.812\\
    \hline
    local OrigCL&0.756&0.907&0.691&0.848\\
    \hline
    MiAUCL&\textbf{0.782}&\textbf{0.917}&\textbf{0.730}&\textbf{0.863}\\
    \hline
  \end{tabular}
\end{table}

\subsection{Visualization.}
\begin{figure}[htbp]
    \centering
    \resizebox{0.5\textwidth}{!}{%
    \begin{tabular}{c}
        \begin{subfigure}[b]{0.23\linewidth}
            \centering
            \includegraphics[width=\linewidth]{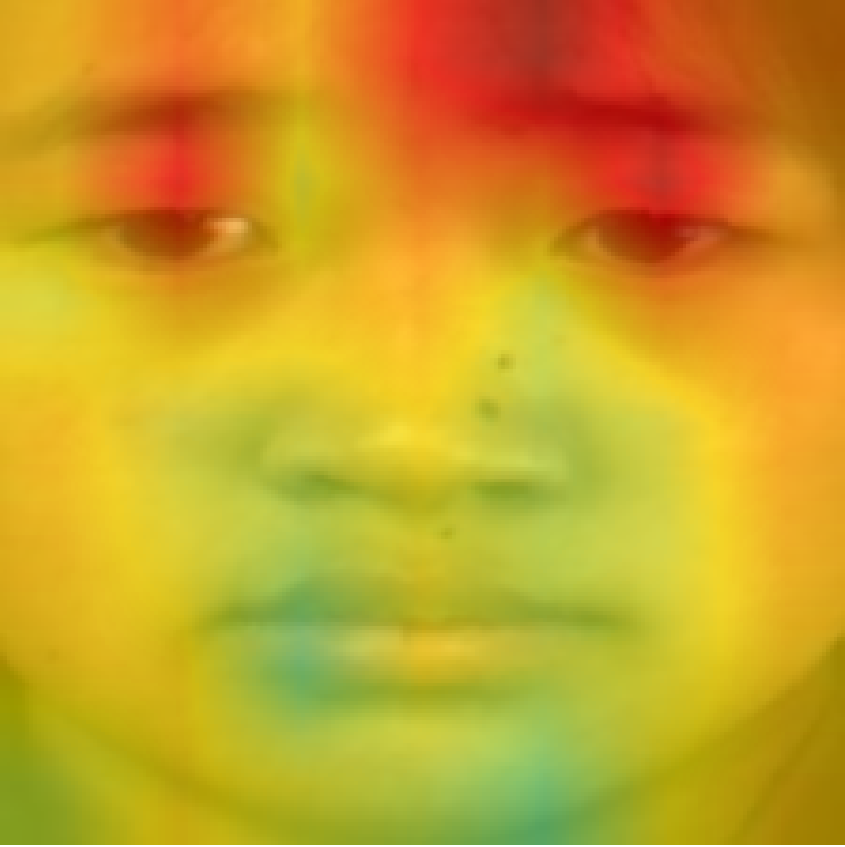}
            \caption{\footnotesize AU1}
            \label{AU-1}
        \end{subfigure} 
        \begin{subfigure}[b]{0.23\linewidth}
            \centering
            \includegraphics[width=\linewidth]{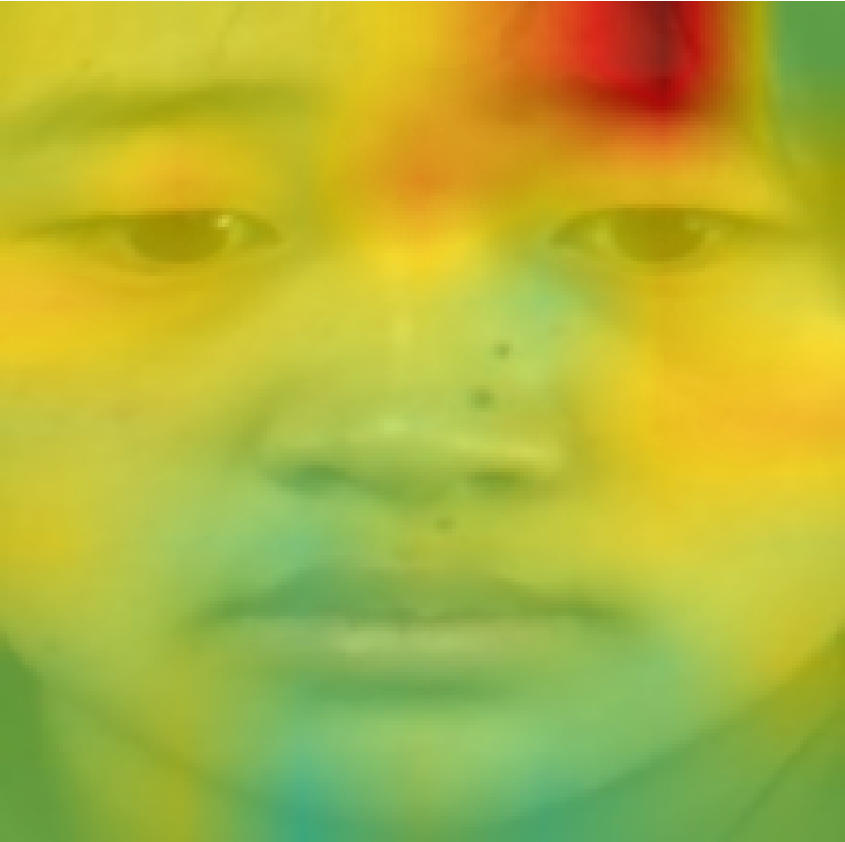}
            \caption{\footnotesize AU1+L2} 
            \label{AU-2}
        \end{subfigure} \\
        \begin{subfigure}[b]{0.23\linewidth}
            \centering
            \includegraphics[width=\linewidth]{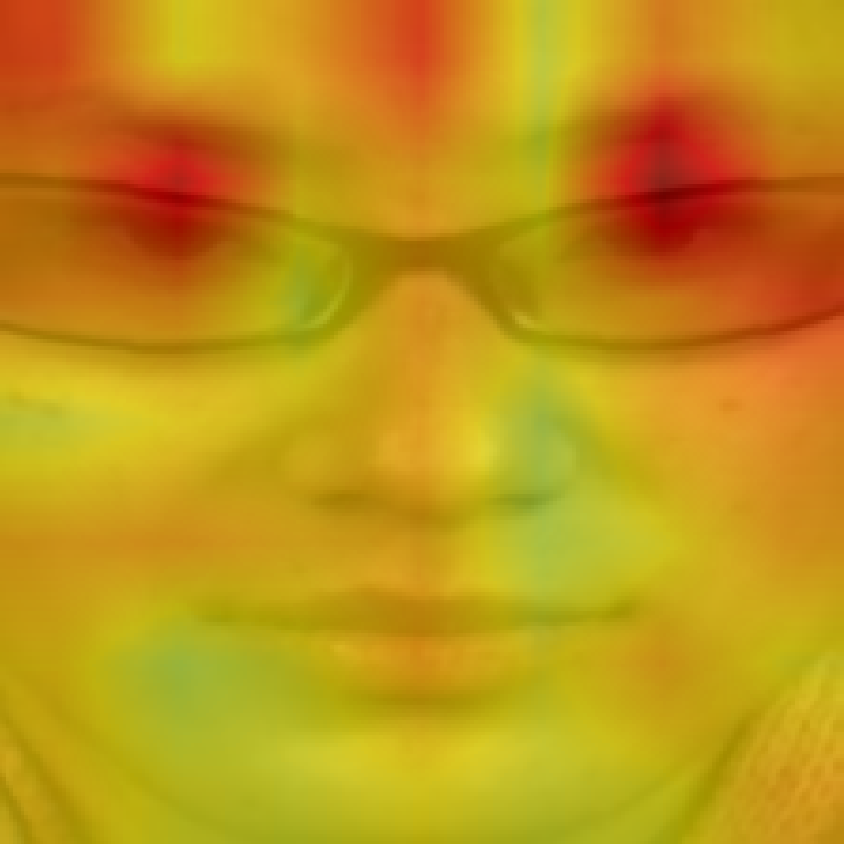}
            \caption{\footnotesize AU4+7}
            \label{AU-3}
        \end{subfigure}  
        \begin{subfigure}[b]{0.23\linewidth}
            \centering
            \includegraphics[width=\linewidth]{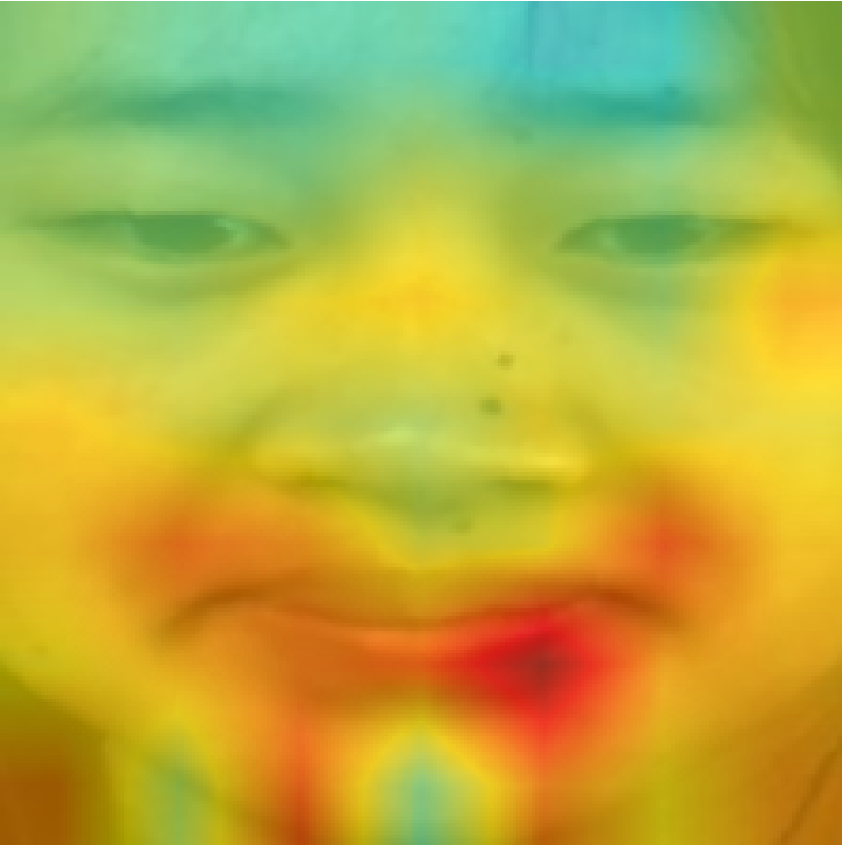}
            \caption{\footnotesize AU12+15}
            \label{AU-4}
        \end{subfigure}
        \begin{subfigure}[b]{0.23\linewidth}
            \centering
            \includegraphics[width=\linewidth]{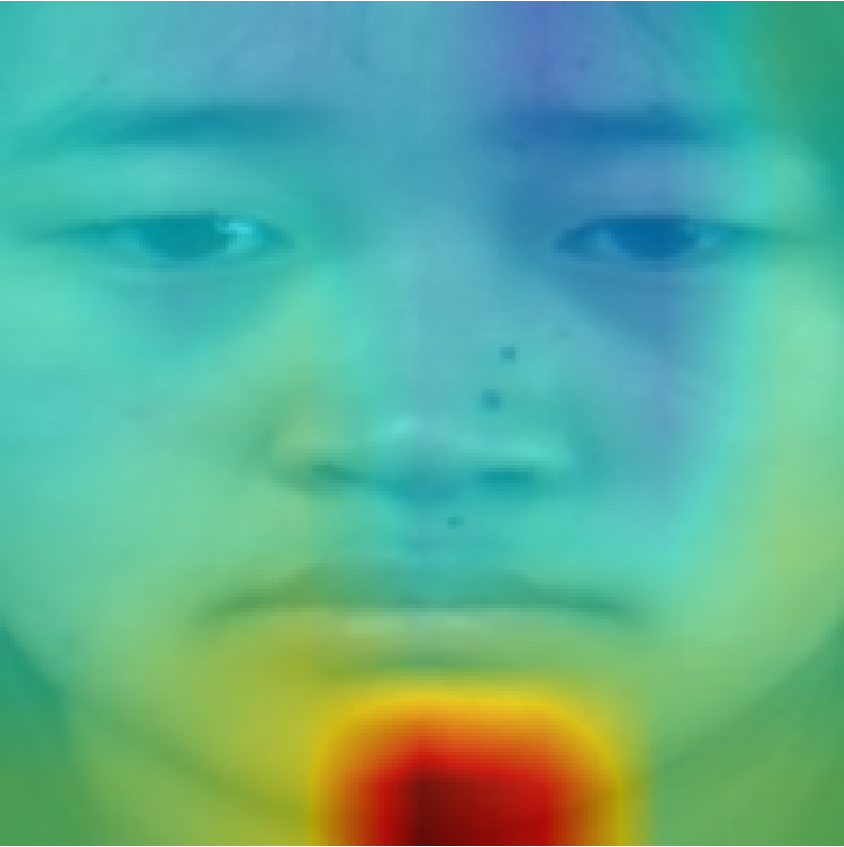}
            \caption{\footnotesize AU17}
            \label{AU-5}
        \end{subfigure}
    \end{tabular}}
    \caption{Attention heatmap visualization results}
    \label{fig:visualize1}
\end{figure}
To further validate the Micro-AU CLIP's interpretability and region-specific sensitivity in AU detection tasks, a visual analysis based on the attention headmap is conducted  as shown in \cref{fig:visualize1}. It turns out that the proposed model mainly focuses on the area where AU is activated. For example, for \cref{AU-1,AU-2}, when AU1 (Inner Brow Raiser) is activated, the model has a higher level of attention in the eyebrow area; for \cref{AU-5}, when AU17 (Chin Lifting Upward) is activated, the model has a higher level of attention in the chin area. 
It demonstrates that Micro-AU CLIP can effectively focus on the relevant regions where AU is activated.

\begin{figure}[htbp]
    \centering
    \resizebox{0.5\textwidth}{!}{%
    \begin{tabular}{c}
        \begin{subfigure}[b]{0.5\linewidth}
            \centering
            \includegraphics[width=\linewidth]{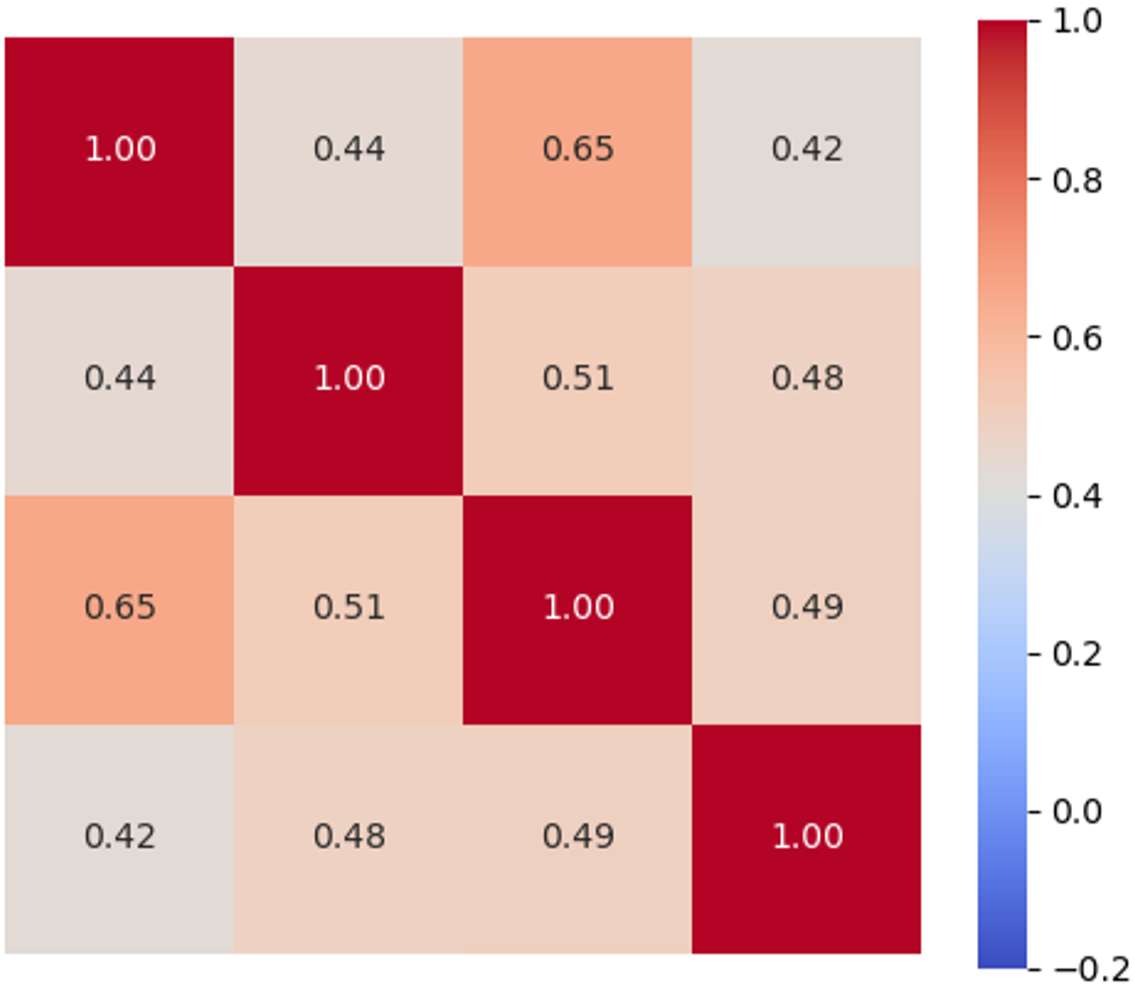}
            \caption{\small Local OrigCL}
            \label{VLO}
        \end{subfigure} 
        \begin{subfigure}[b]{0.5\linewidth}
            \centering
            \includegraphics[width=\linewidth]{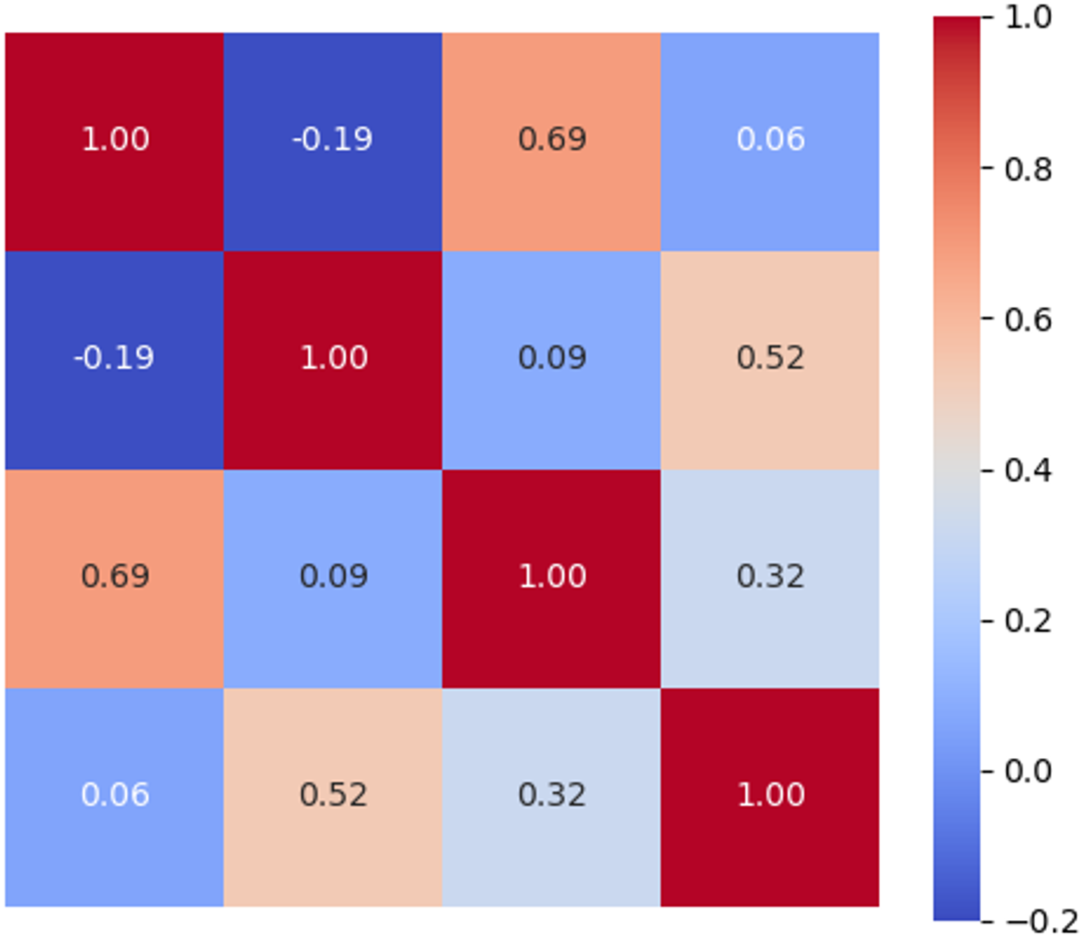}
            \caption{\small MiAUCL}
            \label{VMiAUCL}
        \end{subfigure} 
    \end{tabular}
    }
    \caption{Visualisation of similarity matrix}
    \label{sim}
\end{figure}

Furthermore, as shown in \cref{sim}, we visualize the similarity of visual features between positive (AU label is 1) and negative (AU label is 0) samples under Local OrigCL and MiAUCL. We selected four samples with AU1 label \{1,0,1,0\} and calculated the similarity between their visual features. 
For the similarity (such as row 1 column 2) between a positive sample and a negative sample, the model with Local OrigCL (\cref{VLO}) has a higher similarity than the model with MiAUCL (\cref{VMiAUCL}). The above results indicate that setting the similarity label as a diagonal matrix does not take into account the high similarity between different samples with similar AU labels, while MiAUCL can learn their high similarity. As a result, the proposed MiAUCL can effectively learn the discriminative AU features.
\begin{figure*}[htbp]
    \centering
    \begin{subfigure}[t]{0.32\textwidth}
        \centering
        \includegraphics[width=\linewidth]{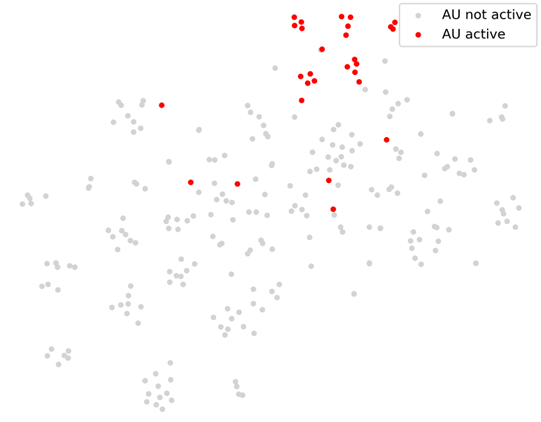}
        \caption{\small Micro-AU CLIP}
    \end{subfigure}
    \hfill
    \begin{subfigure}[t]{0.32\textwidth}
        \centering
        \includegraphics[width=\linewidth]{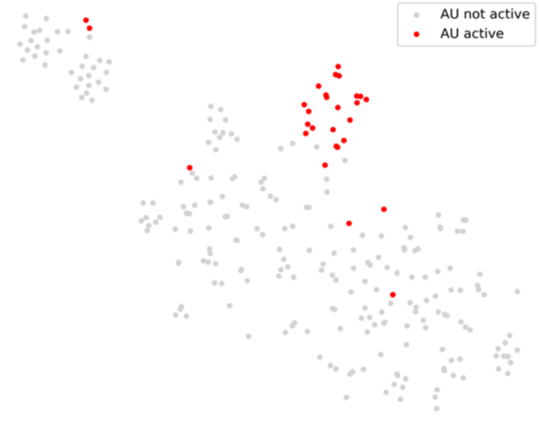}
        \caption{\small Micro-AU CLIP without GSD}
    \end{subfigure}
    \hfill
    \begin{subfigure}[t]{0.32\textwidth}
        \centering
        \includegraphics[width=\linewidth]{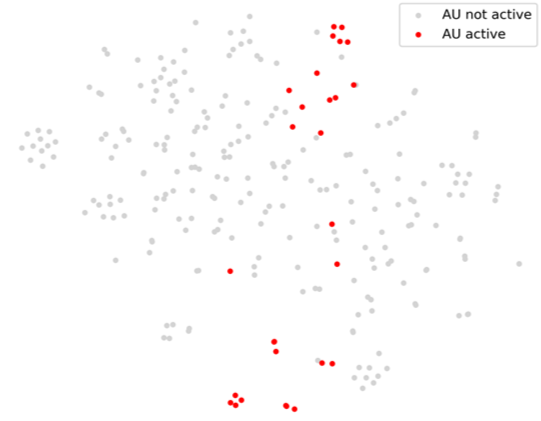}
        \caption{\small Micro-AU CLIP without LSI}
        \label{fig:sub3}
    \end{subfigure}
    \caption{Visualization of the AU1 visual features using t-SNE, with different models.}
    \label{fig:visualize}
\end{figure*}

By visualizing features using t-SNE, we further analyze the effectiveness of the proposed framework. \cref{fig:visualize} shows scatter plots of AU1 features learned by different models. Micro-AU CLIP without LSI performs poorly in both inter-class and intra-class distances. Specifically, there is a significant overlap in feature points between activated and inactivated samples. Compared to it, the other two have advantages by introducing local semantic independence into the construction of the micro-AU detection framework, which demonstrates that learning each micro-AU feature through explicit independent channels is effective. Furthermore, compared to Micro-AU CLIP without GSD, Micro-AU CLIP has better intra-class distance, e.g., a clearly separated feature point area appears in the upper left corner. It demonstrates that introducing GSD can improve the intra-class distance. Overall, the proposed framework can learn micro AU features with smaller intra-class distance and larger inter-class distance through learning each AU features independently and modeling global dependence.

\subsection{Comparing With the Existing Methods}
As shown in \cref{tab:autask}, Micro-AU CLIP is compared to existing methods on CASME II and SAMM. Overall, Micro-AU CLIP outperforms existing methods. Compared to the traditional method and traditional deep-learning models, our method achieves a better performance; Compared to graph model-based ME-GraphAU and Transformer-based AUFormer, the advantage is still maintained. AU-LLM is the SOTA method based on Large Language Model. Although AU-LLM performs better on CASME II, our advantage is very evident on SAMM. Moreover, in terms of the performance difference on the two datasets, Micro-AU CLIP achieves a smaller difference (0.052) compared to AU-LLM (0.195), indicating that our method has better robustness and stability across datasets. Thus, Micro-AU CLIP achieves competitive performance and is more robust.
\begin{table}[t]
\caption{\textsc{F1 Results Compared to Other Methods}}
\label{tab:autask}
\centering
  \begin{tabular}{|c||c||c||c|}\hline

      Methods& Year &CASME II&SAMM\\

\hline
      SP\cite{tung2019similarity} &2019& 0.633 & 0.487\\
      \hline
   LKFSR\cite{zhang2021facial} &2021& 0.506 & 0.455\\
   \hline
  SCA\cite{Li_Huang_Zhao_2021} &2021& 0.668&  0.505\\
  \hline
  ME-GraphAU\cite{luo2022learning}&2022& 0.624 & N/A\\
  \hline
  Resnet18\cite{varanka2023learnable} &2023& 0.773 & 0.573\\
  \hline
  AUFormer\cite{yuan2024auformer} &2024& 0.704 & N/A\\
  \hline
 MG\cite{Zhou2025Objective} &2025& 0.530 & 0.500\\
 \hline
 AU-LLM\cite{liu2025llm} &2025& \textbf{0.814} & 0.619\\
 \hline
   Micro-AU CLIP&2025& 0.782& \textbf{0.730} \\ 
   \hline
  \end{tabular}
\end{table}

\subsection{The Evaluation on MER}
It is worth noting that in the actual micro expression dataset, some AUs have a low frequency of occurrence and were not included in our AU detection task. Therefore, when performing MER, it is necessary to remove the less frequently used AUs based on the standard correspondence between facial AU and ME categories in commonly used AU combinations, and summarize a filtered AU micro expression correspondence. For more detailed information, please refer to the supplementary materials.

As shown in \cref{tabMER}, we evaluated Micro-AU CLIP on the MER task in a zero-shot learning (ZSL) way. Micro-AU CLIP without GDA (w/o GDA) achieves suboptimal results in micro-AU detection task. For this method, our method maintains an advantage in ZSL based MER tasks. Also, compared to existing MER methods based on supervised learning (SL) and weakly-supervised learning (WSL), our method did not achieve optimal performance, but the difference is not significant, even surpassing some latest works with SL, such as CoT on CASME II, TFT on SAMM, and 3CC on both datasets.
The main reason is that supervised learning uses emotional labels for constraining models. Our method explores ZSL-based MER. Compared to these works, our method does not use ME labels, avoiding the impact of label ambiguity on recognition results.

\begin{table}[t]
\caption{\textsc{F1 Results Compared to Other Methods. LW: Learning Way}}
\label{tabMER}
\centering
  \begin{tabular}{|c||c||c||c||c|}
  \hline
    
    Methods& Year &LW&CASME II&SAMM\\
      \hline
 Sup\cite{zhi2022micro} & 2022 & SL& 0.881&0.713 \\ 
 \hline
SME\cite{fan2023selfme}& 2023&SL&\textbf{0.908}&N/A\\ 
\hline
Bert\cite{nguyen2023micron}& 2023&WS& 0.903&N/A\\ 
\hline
3CC\cite{fang2024micro}& 2024&SL&0.723&0.701\\ 
   \hline
    TFT\cite{wang2024two}& 2024&SL&0.907&0.709\\
    \hline
    CoT\cite{yang2025micro}& 2025&SL& 0.802&0.737\\
     \cline{1-5}
 w/o GDA & 2025&ZSL&0.872&0.696\\
 \hline
 Our method& 2025& ZSL&0.889 & \textbf{0.747}\\ \hline
    
  \end{tabular}   
\end{table}

\section{Conclusion}
This paper systematically explored the effectiveness of the framework from local independence to global dependence, and proposes a novel Micro-AU CLIP model that achieves the micro-AU detection and emotion-label-free MER. Firstly, PTA in LSI module aggregates the visual features of each AU to learn its semantic consistency features. Then, GSD module is introduced to transmit contextual information between AU features, thereby modeling the dependencies between AUs and enhancing the features of AUs from a global perspective. Subsequently, we design a MiAUCL loss function to achieve fine-grained alignment between visual semantics and textual semantics, as well as textual guidance for AU visual features; meanwhile, a GD loss is introduced to constrain the dependency modeling between AUs. Through this combination, the framework can simultaneously model local independence and global dependence.

Extensive experiments were conducted, including quantitative analysis and visualization analysis. The experimental results demonstrate that the designed framework, in "independence-to-dependence" pattern, is effective, and Micro-AU CLIP can fully learn fine-grained micro-AU features by modeling local semantic independence and global semantic dependence. Also, the ablation analysis demonstrates the rationality and effectiveness of LSI, GSD, PTA, GDA, MiAUC Loss and GD Loss for the micro-AU detection task. Finally, compared with SOTA methods, the proposed method achieves competitive performance, and a more robust result across datasets. Furthermore, Micro-AU CLIP can effectively be applied to MER with emotion-label-free and achieve a competitive performance. 


\section*{Acknowledgments}
The authors would like to thank all collaborators involved in this work for their valuable discussions and contributions. The authors also sincerely appreciate the constructive suggestions from the anonymous reviewers, which helped improve the quality of this paper. In addition, we thank the providers of the public micro-expression datasets used in this study for making their resources available to the research community.




 


\bibliographystyle{IEEEtran}
\bibliography{main}

@String(ECCV= {Eur. Conf. Comput. Vis.})

@String(ICASSP=	{ICASSP})

@String(ECCV  = {ECCV})

@inproceedings{wei2025multi,
  title={Multi-Information Hierarchical Fusion Transformer with Local Alignment and Global Correlation for Micro-Expression Recognition},
  author={Wei, Jinsheng and Sun, Jialiang and Lu, Guanming and Yan, Jingjie and Zhang, Dong},
  booktitle={Proceedings of the 33rd ACM International Conference on Multimedia},
  pages={5873--5882},
  year={2025}
}

@article{wei2022learning,
  title={Learning two groups of discriminative features for micro-expression recognition},
  author={Wei, Jinsheng and Lu, Guanming and Yan, Jingjie and Zong, Yuan},
  journal={Neurocomputing},
  volume={479},
  pages={22--36},
  year={2022},
  publisher={Elsevier}
}

@inproceedings{fan2023selfme,
  title={SelfME: Self-supervised motion learning for micro-expression recognition},
  author={Fan, Xinqi and Chen, Xueli and Jiang, Mingjie and Shahid, Ali Raza and Yan, Hong},
  booktitle={Proceedings of the IEEE/CVF conference on computer vision and pattern recognition},
  pages={13834--13843},
  year={2023}
}

@article{wei2023prior,
  title={Prior information based decomposition and reconstruction learning for micro-expression recognition},
  author={Wei, Jinsheng and Chen, Haoyu and Lu, Guanming and Yan, Jingjie and Xie, Yue and Zhao, Guoying},
  journal={IEICE TRANSACTIONS on Information and Systems},
  volume={106},
  number={10},
  pages={1752--1756},
  year={2023},
  publisher={The Institute of Electronics, Information and Communication Engineers}
}

@article{Shao_Liu_Cai_Ma_2021,   title={JÂA-Net: Joint Facial Action Unit Detection and Face Alignment Via Adaptive Attention},  url={http://dx.doi.org/10.1007/s11263-020-01378-z},  DOI={10.1007/s11263-020-01378-z},  journal={International Journal of Computer Vision},  author={Shao, Zhiwen and Liu, Zhilei and Cai, Jianfei and Ma, Lizhuang},  year={2021},  month={Feb},  pages={321–340},  language={en-US}  }

@article{li2018eac,
  title={Eac-net: Deep nets with enhancing and cropping for facial action unit detection},
  author={Li, Wei and Abtahi, Farnaz and Zhu, Zhigang and Yin, Lijun},
  journal={IEEE transactions on pattern analysis and machine intelligence},
  volume={40},
  number={11},
  pages={2583--2596},
  year={2018},
  publisher={IEEE}
}

@article{shao2019facial,
  title={Facial action unit detection using attention and relation learning},
  author={Shao, Zhiwen and Liu, Zhilei and Cai, Jianfei and Wu, Yunsheng and Ma, Lizhuang},
  journal={IEEE transactions on affective computing},
  volume={13},
  number={3},
  pages={1274--1289},
  year={2019},
  publisher={IEEE}
}

@inproceedings{radford2021learning,
  title={Learning transferable visual models from natural language supervision},
  author={Radford, Alec and Kim, Jong Wook and Hallacy, Chris and Ramesh, Aditya and Goh, Gabriel and Agarwal, Sandhini and Sastry, Girish and Askell, Amanda and Mishkin, Pamela and Clark, Jack and others},
  booktitle={International conference on machine learning},
  pages={8748--8763},
  year={2021},
  organization={PmLR}
}

@article{wei2023geometric,
  title={Geometric graph representation with learnable graph structure and adaptive au constraint for micro-expression recognition},
  author={Wei, Jinsheng and Peng, Wei and Lu, Guanming and Li, Yante and Yan, Jingjie and Zhao, Guoying},
  journal={IEEE Transactions on Affective Computing},
  volume={15},
  number={3},
  pages={1343--1357},
  year={2023},
  publisher={IEEE}
}

@article{li2023con,
  title={Contrastive learning of person-independent representations for facial action unit detection},
  author={Li, Yong and Shan, Shiguang},
  journal={IEEE Transactions on Image Processing},
  volume={32},
  pages={3212--3225},
  year={2023},
  publisher={IEEE}
}

@inproceedings{li2021intra,
  title={Intra-and inter-contrastive learning for micro-expression action unit detection},
  author={Li, Yante and Zhao, Guoying},
  booktitle={Proceedings of the 2021 International Conference on Multimodal Interaction},
  pages={702--706},
  year={2021}
}

@article{liu2025mer,
  title={MER-CLIP: AU-Guided Vision-Language Alignment for Micro-Expression Recognition},
  author={Liu, Shifeng and Mao, Xinglong and Zhao, Sirui and Li, Peiming and Xu, Tong and Chen, Enhong},
  journal={IEEE Transactions on Affective Computing},
  year={2025},
  publisher={IEEE}
}

@article{gong2023meta,
  title={Meta-MMFNet: Meta-learning-based multi-model fusion network for micro-expression recognition},
  author={Gong, Wenjuan and Zhang, Yue and Wang, Wei and Cheng, Peng and Gonzalez, Jordi},
  journal={ACM Transactions on Multimedia Computing, Communications and Applications},
  volume={20},
  number={2},
  pages={1--20},
  year={2023},
  publisher={ACM New York, NY}
}

@article{cai2024mfdan,
  title={Mfdan: Multi-level flow-driven attention network for micro-expression recognition},
  author={Cai, Wenhao and Zhao, Junli and Yi, Ran and Yu, Minjing and Duan, Fuqing and Pan, Zhenkuan and Liu, Yong-Jin},
  journal={IEEE Transactions on Circuits and Systems for Video Technology},
  year={2024},
  publisher={IEEE}
}

@inproceedings{qi2024multimodal,
  title={Multimodal emotion recognition with vision-language prompting and modality dropout},
  author={Qi, Anbin and Liu, Zhongliang and Zhou, Xinyong and Xiao, Jinba and Zhang, Fengrun and Gan, Qi and Tao, Ming and Zhang, Gaozheng and Zhang, Lu},
  booktitle={Proceedings of the 2nd International Workshop on Multimodal and Responsible Affective Computing},
  pages={49--53},
  year={2024}
}

@inproceedings{nguyen2023micron,
  title={Micron-bert: Bert-based facial micro-expression recognition},
  author={Nguyen, Xuan-Bac and Duong, Chi Nhan and Li, Xin and Gauch, Susan and Seo, Han-Seok and Luu, Khoa},
  booktitle={Proceedings of the ieee/cvf conference on computer vision and pattern recognition},
  pages={1482--1492},
  year={2023}
}

@inproceedings{varanka2023learnable,
  title={Learnable Eulerian dynamics for micro-expression action unit detection},
  author={Varanka, Tuomas and Peng, Wei and Zhao, Guoying},
  booktitle={Scandinavian Conference on Image Analysis},
  pages={385--400},
  year={2023},
  organization={Springer}
}

@article{Zhou2025Objective, 
author = {Ling Zhou and Qirong Mao and Ming Dong},
title = {Objective Class-Based Micro-Expression Recognition Through Simultaneous Action Unit Detection and Feature Aggregation},
year = {2025},
journal = {Tsinghua Science and Technology},
volume = {30},
number = {5},
pages = {2114-2132},
url = {https://www.sciopen.com/article/10.26599/TST.2024.9010095},
doi = {10.26599/TST.2024.9010095}
}

@article{Li_Huang_Zhao_2021,   title={Micro-expression action unit detection with spatial and channel attention},  url={http://dx.doi.org/10.1016/j.neucom.2021.01.032},  DOI={10.1016/j.neucom.2021.01.032},  journal={Neurocomputing},  author={Li, Yante and Huang, Xiaohua and Zhao, Guoying},  year={2021},  month={May},  pages={221–231},  language={en-US}  }

@article{Yan_Li_Wang_Zhao_Liu_Chen_Fu_2014,  
 title={CASME II: An Improved Spontaneous Micro-Expression Database and the Baseline Evaluation}, 
 url={http://dx.doi.org/10.1371/journal.pone.0086041}, 
 DOI={10.1371/journal.pone.0086041}, 
 journal={PLoS ONE}, 
 author={Yan, Wen-Jing and Li, Xiaobai and Wang, Su-Jing and Zhao, Guoying and Liu, Yong-Jin and Chen, Yu-Hsin and Fu, Xiaolan}, 
 year={2014}, 
 month={Jan}, 
 pages={e86041}, 
 language={en-US} 
 }

@article{davison2016samm,
  title={Samm: A spontaneous micro-facial movement dataset},
  author={Davison, Adrian K and Lansley, Cliff and Costen, Nicholas and Tan, Kevin and Yap, Moi Hoon},
  journal={IEEE transactions on affective computing},
  volume={9},
  number={1},
  pages={116--129},
  year={2016},
  publisher={IEEE}
}

@inproceedings{zhang2021facial,
  title={Facial action unit detection with local key facial sub-region based multi-label classification for micro-expression analysis},
  author={Zhang, Liangfei and Arandjelovic, Ognjen and Hong, Xiaopeng},
  booktitle={Proceedings of the 1st workshop on facial micro-expression: advanced techniques for facial expressions generation and spotting},
  pages={11--18},
  year={2021}
}

@inproceedings{li2021micro,
  title={Micro-expression action unit detection with dual-view attentive similarity-preserving knowledge distillation},
  author={Li, Yante and Peng, Wei and Zhao, Guoying},
  booktitle={2021 16th IEEE International Conference on Automatic Face and Gesture Recognition (FG 2021)},
  pages={01--08},
  year={2021},
  organization={IEEE}
}

@article{zhang2022short,
  title={Short and long range relation based spatio-temporal transformer for micro-expression recognition},
  author={Zhang, Liangfei and Hong, Xiaopeng and Arandjelovi{\'c}, Ognjen and Zhao, Guoying},
  journal={IEEE Transactions on Affective Computing},
  volume={13},
  number={4},
  pages={1973--1985},
  year={2022},
  publisher={IEEE}}

@article{liong2018less,
  title={Less is more: Micro-expression recognition from video using apex frame},
  author={Liong, Sze-Teng and See, John and Wong, KokSheik and Phan, Raphael C-W},
  journal={Signal Processing: Image Communication},
  volume={62},
  pages={82--92},
  year={2018},
  publisher={Elsevier}
}

@inproceedings{tung2019similarity,
  title={Similarity-preserving knowledge distillation},
  author={Tung, Frederick and Mori, Greg},
  booktitle={Proceedings of the IEEE/CVF international conference on computer vision},
  pages={1365--1374},
  year={2019}
}

@inproceedings{luo2022learning,
  title={Learning Multi-dimensional Edge Feature-based AU Relation Graph for Facial Action Unit Recognition},
  author={Luo, Cheng and Song, Siyang and Xie, Weicheng and Shen, Linlin and Gunes, Hatice},
  year={2022},
  organization={International Joint Conferences on Artificial Intelligence Organization}
}

@inproceedings{yuan2024auformer,
  title={Auformer: Vision transformers are parameter-efficient facial action unit detectors},
  author={Yuan, Kaishen and Yu, Zitong and Liu, Xin and Xie, Weicheng and Yue, Huanjing and Yang, Jingyu},
  booktitle={European Conference on Computer Vision},
  pages={427--445},
  year={2024},
  organization={Springer}
}

@inproceedings{fang2024micro,
  title={Micro-Expression Recognition by 3D-CNN in Combination with GRU},
  author={Fang, Chun-Ting and Liu, Tsung-Jung and Liu, Kuan-Hsien},
  booktitle={2024 International Conference on Consumer Electronics-Taiwan (ICCE-Taiwan)},
  pages={95--96},
  year={2024},
  organization={IEEE}
}

@article{wang2024two,
  title={Two-level spatio-temporal feature fused two-stream network for micro-expression recognition},
  author={Wang, Zebiao and Yang, Mingyu and Jiao, Qingbin and Xu, Liang and Han, Bing and Li, Yuhang and Tan, Xin},
  journal={Sensors},
  volume={24},
  number={5},
  pages={1574},
  year={2024},
  publisher={MDPI}
}

@article{yang2025micro,
  title={Micro-expression recognition based on contextual transformer networks},
  author={Yang, Jun and Wu, Zilu and Wu, Renbiao},
  journal={The Visual Computer},
  volume={41},
  number={3},
  pages={1527--1541},
  year={2025},
  publisher={Springer}
}

@article{auflem2022facing,
  title={Facing the facs—using ai to evaluate and control facial action units in humanoid robot face development},
  author={Auflem, Marius and Kohtala, Sampsa and Jung, Malte and Steinert, Martin},
  journal={Frontiers in Robotics and AI},
  volume={9},
  pages={887645},
  year={2022},
  publisher={Frontiers Media SA}
}

@article{frank1997ability,
  title={The ability to detect deceit generalizes across different types of high-stake lies.},
  author={Frank, Mark G and Ekman, Paul},
  journal={Journal of personality and social psychology},
  volume={72},
  number={6},
  pages={1429},
  year={1997},
  publisher={American Psychological Association}
}

@article{huang2016spontaneous,
  title={Spontaneous facial micro-expression analysis using spatiotemporal completed local quantized patterns},
  author={Huang, Xiaohua and Zhao, Guoying and Hong, Xiaopeng and Zheng, Wenming and Pietik{\"a}inen, Matti},
  journal={Neurocomputing},
  volume={175},
  pages={564--578},
  year={2016},
  publisher={Elsevier}
}

@article{Ekman_Friesen_1978,  
 title={Facial action coding system: a technique for the measurement of facial movement}, 
 author={Ekman, Paul and Friesen, WallaceV.}, 
 year={1978}, 
 month={Jan}, 
 language={en-US} 
 }

@inproceedings{wang2024micro,
  title={Micro-expression recognition by fusing action unit detection and Spatio-temporal features},
  author={Wang, Lei and Huang, Pinyi and Cai, Wangyang and Liu, Xiyao},
  booktitle={ICASSP 2024-2024 IEEE International Conference on Acoustics, Speech and Signal Processing (ICASSP)},
  pages={5595--5599},
  year={2024},
  organization={IEEE}
}

@article{liu2025llm,
  title={AU-LLM: Micro-Expression Action Unit Detection via Enhanced LLM-Based Feature Fusion},
  author={Liu, Zhishu and Yuan, Kaishen and Zhao, Bo and Xu, Yong and Yu, Zitong},
  journal={arXiv preprint arXiv:2507.21778},
  year={2025}
}

@article{zhi2022micro,
  title={Micro-expression recognition with supervised contrastive learning},
  author={Zhi, Ruicong and Hu, Jing and Wan, Fei},
  journal={Pattern Recognition Letters},
  volume={163},
  pages={25--31},
  year={2022},
  publisher={Elsevier}
}

@article{zhao2023clip,
  title={CLIP in Medical Imaging: A Comprehensive Survey},
  author={Zhao, Zihao and Liu, Yuxiao and Wu, Han and Li, Yonghao and Wang, Sheng and Teng, Lin and Liu, Disheng and Cui, Zhiming and Wang, Qian and Shen, Dinggang},
  journal={CoRR},
  year={2023}
}

@inproceedings{zhang2025clip,
  title={Clip-moe: Towards building mixture of experts for clip with diversified multiplet upcycling},
  author={Zhang, Jihai and Qu, Xiaoye and Zhu, Tong and Cheng, Yu},
  booktitle={Proceedings of the 2025 Conference on Empirical Methods in Natural Language Processing},
  pages={5406--5419},
  year={2025}
}

@article{zhou2023micro,
  title={Micro-expression action unit recognition based on dynamic image and spatial pyramid.},
  author={Zhou, Guanqun and Yuan, Shusen and Xing, Hongbo and Jiang, Youjun and Geng, Pinyong and Cao, Yewen and Ben, Xianye},
  journal={Journal of Supercomputing},
  volume={79},
  number={17},
  year={2023}
}

@inproceedings{oh2018learning,
  title={Learning-based video motion magnification},
  author={Oh, Tae-Hyun and Jaroensri, Ronnachai and Kim, Changil and Elgharib, Mohamed and Durand, Fr'edo and Freeman, William T and Matusik, Wojciech},
  booktitle={Proceedings of the European conference on computer vision (ECCV)},
  pages={633--648},
  year={2018}
}

@article{van2021bayesian,
  title={Bayesian statistics and modelling},
  author={Van de Schoot, Rens and Depaoli, Sarah and King, Ruth and Kramer, Bianca and M{\"a}rtens, Kaspar and Tadesse, Mahlet G and Vannucci, Marina and Gelman, Andrew and Veen, Duco and Willemsen, Joukje and others},
  journal={Nature Reviews Methods Primers},
  volume={1},
  number={1},
  pages={1},
  year={2021},
  publisher={Nature Publishing Group UK London}
}

\vspace{11pt}


\vspace{11pt}


\vfill

\end{document}